\documentclass[runningheads]{llncs}
\usepackage{graphicx}
\usepackage{amsmath,amssymb} 
\usepackage{color}

\usepackage{stmaryrd}
\usepackage{algorithm}
\usepackage{algorithmicx}
\usepackage[noend]{algpseudocode}
\usepackage{booktabs}
\usepackage{caption}
\usepackage{subcaption}
\begin{document}

\newcommand{\point}{
    \raise0.7ex\hbox{.}
    }


\pagestyle{headings}

\mainmatter

\title{Joint Training of Generic CNN-CRF Models with Stochastic Optimization} 

\titlerunning{Joint Training of Generic CNN-CRF Models with Stochastic Optimization} 

\authorrunning{A.~Kirillov, D.~Schlesinger, S.~Zheng, B.~Savchynskyy,  P.H.S.~Torr, C.~Rother} 

\author{A.~Kirillov$^1$, D.~Schlesinger$^1$, S.~Zheng$^2$, B.~Savchynskyy$^1$,  P.H.S.~Torr$^2$, C.~Rother$^1$} 
\institute{$^1$Dresden University of Technology, $^2$University of Oxford} 

\maketitle

\begin{abstract}
We propose a new CNN-CRF end-to-end learning framework, which is based on joint stochastic optimization with respect to both Convolutional Neural Network (CNN) and Conditional Random Field (CRF) parameters. While stochastic gradient descent is a standard technique for CNN training, it was not used for joint models so far. We show that our learning method is (i) general, i.e.~it applies to arbitrary CNN and CRF architectures and potential functions; (ii) scalable, i.e.~it has a low memory footprint and straightforwardly parallelizes on GPUs; (iii) easy in implementation. Additionally, the unified CNN-CRF optimization approach simplifies a potential hardware implementation. We empirically evaluate our method on the task of semantic labeling of body parts in depth images and show that it compares favorably to competing techniques.
\end{abstract}


\section{Introduction}

Deep learning have tremendous success since a few years in many areas of computational science. In computer vision, Convolutional Neural Networks (CNNs) are successfully used in a wide range of applications -- from low-level vision, like segmentation and optical flow, to high-level vision, like scene understanding and semantic segmentation. For instance in the VOC2012 object segmentation challenge\footnote{http://host.robots.ox.ac.uk:8080/leaderboard} the use of CNNs has pushed the quality score by around 28\% (from around 50\% to currently around 78\% \cite{LinSRH15}). The main contribution of CNNs is their ability to adaptively fine-tune millions of features to achieve best performance for the task at hand. However, CNNs have also their shortcomings. One limitation is that often a large corpus of labeled training images is necessary. Secondly, it is difficult to incorporate prior knowledge into the CNN architecture.  In contrast, graphical models like Conditional Random Fields (CRFs) \cite{lafferty2001conditional} overcome these two limitations.  CRFs have been used to model geometric properties, such as object shape, spatial relationship between objects, global properties like object connectivity, and many others. Furthermore, CRFs designed based on e.g.~physical properties are able to achieve good results even with few training images. For these reasons, a recent trend has been to explore the combination of these two modeling paradigms by using a CRF, whose factors are dependent on a CNN. By doing so, CRFs are able to use the incredible power of CNNs, to fine-tune model features. On the other hand, CNNs can more easily capture global properties such as object shape and contextual information. 
The study of this fruitful combination (sometimes called ``deep structured models''~\cite{ChenSYU15}) is the main focus of our work. We propose a generic joint learning framework for the combined CNN-CRF models, based on a sampling technique and a stochastic gradient optimization.

\subsubsection{Related work.} 

The idea of making CRF models more powerful by allowing factors to depend on many parameters has been explored extensively over the last decade. One example is the Decision Tree Field approach \cite{nowozin2011decision} where factors are dependent on Decision Trees. In this work, we are interested in making the factors dependent on CNNs. Note that one advantage of CNNs over Decision Trees is that CNNs learn the appropriate features for the task at hand, while Decision Trees, as many other classifiers, only combine and select from a pool of simple features, see e.g.~\cite{sethi1990entropy,richmond2015relating} for a discussion on the relationship between CNNs and Decision Trees. We now describe the most relevant works that combine CNNs and CRFs in the context of semantic segmentation, as one of the largest application areas of this type of models. The framework we propose in this work is also evaluated in a similar scenario, although its theoretical basis is application-independent.

Since CNNs have been used for semantic segmentation, this field has made a big leap forward, see e.g.~\cite{long2014fully,lecun/pami2013}. Recently, the advantages of additionally integrating a CRF model have given a further boost in performance, as demonstrated by many works. To the extent that the work \cite{LinSRH15} is currently leading the VOC2012 object segmentation challenge, as discussed below. In \cite{chen2014semantic} a fully connected Gaussian CRF model \cite{krahenbuhl2012efficient} was used, where the respective unaries were supplied by a CNN. The CRF inference was done with a Mean Field approximation. This separate training procedure was recently improved in \cite{ZhengJRVSDHT15} with an end-to-end learning algorithm. To achieve this, they represent the Mean Field iterations as a Recurrent Neural Network. The same idea was published in \cite{schwing/arxiv2015}. In \cite{krahenbuhl2012efficient}, the Mean Field iterations were made efficient by using a so-called permutohedral lattice approximation \cite{adams2010fast} for Gaussian filters. However, this approach allows for a special class of pairwise potentials only. Besides the approaches \cite{ZhengJRVSDHT15} and \cite{schwing/arxiv2015}, there are many other works that consider the idea of backpropagation with a so-called unrolled CRF-inference scheme, such as  \cite{Domke_pami2013,Kiefel_eccv2014,Barbu_tip2009,Ross_cvpr2011,Stoyanov_aistats2011,lecun/nips2014,Ziwei/iccv2015}. These inference steps mostly correspond to message passing operations of e.g.~Mean Field updates or Belief Propagation. However the number of inference iterations in such learning schemes remains their critical parameter: too few iterations lead to a quality deterioration, whereas more iterations slow down the whole learning procedure.
 
Likelihood maximization is NP-hard for CRFs, which implies that it is also NP-hard for joint CNN-CRF models. To avoid this problem, {\em piece-wise} learning~\cite{sutton05piecewise} was used in~\cite{LinSRH15}. Instead of likelihood maximization a surrogate loss is considered which can be minimized efficiently. However, there are no guarantees that minimization of the surrogate loss will lead to maximization of the true likelihood. On the positive side, the method shows good practical results and leads the VOC2012 object segmentation competition at the moment.

Another likelihood approximation, which is based on fractional entropy and a message passing based inference, was proposed in~\cite{ChenSYU15}. However, there is no clear evidence that the fractional entropy always leads to tight likelihood approximations. Another point relates to the memory footprint of the method. To avoid the time consuming, full inference, authors of~\cite{ChenSYU15} interleave gradient steps w.r.t.~the CNN parameters and minimization over the dual variables of the LP-relaxation of the CRF. This allows to solve the issue with a small number of inference iterations comparing to the unrolled inference schemes. However, it requires to store current values of the dual variables for {\em each} element of a training set. The number of the dual variables is proportional to the number of labels in the used CRF as well as to the number of its pairwise factors. Therefore, the size of such a storage can significantly exceed the size required for the training set itself. We will discuss this point in more details in Section~\ref{sec:competing-approaches}. 

\subsubsection{Contribution.}  Inspired by the contrastive divergence approach~\cite{richard2004multiscale}, we propose a {\em generic joint maximum likelihood learning framework} for the combined CNN-CRF models. In this context, ``{\em generic}''  means that (i) factors in our CRF are of a non-parametric form, in contrast to e.g.~\cite{ZhengJRVSDHT15}, where Gaussian pairwise potentials are considered; and (b) we maximize the likelihood itself instead of its approximations. Our framework is based on a sampling technique and stochastic gradient updates w.r.t.~both CNN and CRF parameters. To avoid the time consuming, full inference we interleave sampling-based inference steps with CNN parameters updates. In terms of the memory overhead, our method stores only a single (current) labeling for each element of the training set during learning. This requires less memory than the training set itself. Our method is efficient, scalable and highly parallelizable with a low memory footprint, which makes it an ideal candidate for a GPU-based implementation.

We show the efficiency of our approach on the task of semantic labeling of body parts in depth images.


\section{Preliminaries}

\subsubsection{Conditional Random Fields.}
Let $\vec{y} = (y_1, \ldots, y_N)$ be a random {\em state} vector, where each coordinate is a random variable $y_i$ that takes its values from a finite set ${\cal Y}_i = \{1, \ldots, |{\cal Y}_i|\}$. Therefore $\vec{y}\in {\cal Y} := \prod_{i=1}^{N} {\cal Y}_i$, where $\prod$ stands for a Cartesian product. Let $\vec{x}$ be {\em an observation} vector, taking its values in some set ${\cal X}$. The {\em energy function} $E\colon {\cal Y} \times {\cal X} \times \mathbb{R}^m \rightarrow \mathbb{R}$ assigns a score $E(\vec{y}, \vec{x}, \vec{\theta})$ to a pair $(\vec{y},\vec{x})$ of a state and an observation vector and is parametrized by a {\em parameter} vector  $\vec{\theta} \in \mathbb{R}^m$. An exponential posterior distribution related to the energy $E$ reads
\begin{align}\label{eq:posterior}
	p(\vec{y} | \vec{x}, \vec{\theta}) = \frac{1}{Z(\vec{x}, \vec{\theta})}\exp(-E(\vec{y}, \vec{x}, \vec{\theta})) \,.
\end{align}
Here $Z(\vec{x}, \vec{\theta})$ is a partition function, defined as
\begin{align}
	Z(\vec{x}, \vec{\theta}) = \sum_{\vec{y} \in {\cal Y}} \exp(-E(\vec{y}, \vec{x}, \vec{\theta})) \,.
\end{align}

Let $I=1,...,N$ be a set of {\em variable indexes} and $2^I$ denote its powerset. Let also ${\cal Y}_A$ stand for the set $\prod_{i\in A}{\cal Y}_i$ for any $A \subseteq I$.
In CRFs, the energy function $E$ can be represented as a sum of its components depending on the subsets of variables $\vec{y}_f\in{\cal Y}_f$, $f\subset I$:
\begin{align}\label{eq:crf}
	E(\vec{y}, \vec{x}, \vec{\theta}) = \sum_{f \in {\cal F}\subset 2^{I}} \psi_f(\vec{y}_f, \vec{x}, \vec{\theta}) \,.
\end{align}
The functions $\psi_f\colon {\cal Y}_f\times{\cal X}\times{\mathbb R}^m\to{\mathbb R}$ are usually called {\em potentials}. For example, in~\cite{chen2014semantic,ZhengJRVSDHT15} only CRFs with {\em unary} and {\em pairwise} potentials are considered, i.e.~$|f| \le 2$ for any $f \in {\cal F}$.

In what follows, we will assume that each $\psi_f$ is potentially a non-linear function of $\vec\theta$ and $\vec x$. It can be defined by e.g.~a CNN with the input~$\vec{x}$ and weights~$\vec{\theta}$.

\paragraph{Inference} is a process of estimating the state vector $\vec y$ for an observation $\vec x$. There are several inference criteria, see e.g.~\cite{wainwright2008graphical}. In this work we will stick to the so called {\em maximum posterior marginals}, or shortly {\em max-marginal} inference
\begin{align}\label{eq:mpm}
	y_i^* = \arg\max_{y_i \in {\cal Y}_i} p(y_i | \vec{x}, \vec{\theta}): = \arg\max_{y_i \in {\cal Y}_i} \sum_{(\vec{y'} \in {\cal Y} \colon y'_i=y_i) } p(\vec{y'} | \vec{x}, \vec{\theta})  \qquad \text{ for all $i$} \,.
\end{align}
Though maximization in~\eqref{eq:mpm} can be done directly due to the typically small size of the sets ${\cal Y}_i$, computing the marginals $p(y_i | \vec{x}, \vec{\theta})$ is NP-hard in general. Summation in~\eqref{eq:mpm} can not be performed directly due to the exponential size of the set ${\cal Y}$. In our framework we approximate the marginals with Gibbs sampling~\cite{Gemans}. The corresponding estimates converge to the true marginals in the limit. We detail this procedure in Section~\ref{sec:sampling-based-learning}.

\paragraph{Learning.} Given a training set $\{ (\vec{x}^d, \vec{y}^d) \in ({\cal X} \times {\cal Y})\}_{d=1}^{D}$, we consider the maximum likelihood learning criterion for estimating $\vec\theta$:
\begin{align}\label{eq:likelihood}
	\arg\max_{\vec{\theta} \in \mathbb{R}^m} \sum_{d = 1}^{D} \log p(\vec{y}^d | \vec{x}^d, \vec{\theta}) = \arg\max_{\vec{\theta} \in \mathbb{R}^m} \sum_{d = 1}^{D} \left[ -E(\vec{y}^d, \vec{x}^d, \vec{\theta}) - \log Z(\vec{x}^d, \vec{\theta}) \right] \,.
\end{align}
Since a (stochastic) gradient descent is used for CNN training, we stick to it for estimating~\eqref{eq:likelihood} as well. The gradient of the objective reads:
\begin{align}\label{eq:likelihood_gradient}
&\frac{\partial \sum_{d = 1}^{D} \log p(\vec{y}^d | \vec{x}^d, \vec{\theta})}{\partial\vec{\theta}}  = \sum_{d = 1}^{D} \left[ -\frac{\partial E(\vec{y}^d, \vec{x}^d, \vec{\theta})}{\partial \vec{\theta}} - \frac{\partial \log Z(\vec{x}^d, \vec{\theta})}{\partial\vec{\theta}} \right] \nonumber \\
&\quad= \sum_{d = 1}^{D} \left[ - \frac{\partial E(\vec{y}^d, \vec{x}^d, \vec{\theta})}{\partial\vec{\theta}} - \frac{1}{Z(\vec{x}^d, \vec{\theta})}\frac{\partial \sum_{\vec{y} \in {\cal Y}} \exp(- E(\vec{y}, \vec{x}^d, \vec{\theta}))}{\partial\vec{\theta}} \right] \nonumber \\
&\quad= \sum_{d = 1}^{D} \left[ - \frac{\partial E(\vec{y}^d, \vec{x}^d, \vec{\theta})}{\partial\vec{\theta}} + \sum_{\vec{y} \in {\cal Y}}\frac{\exp(- E(\vec{y}, \vec{x}^d, \vec{\theta}))}{Z(\vec{x}^d, \vec{\theta})}\frac{\partial E(\vec{y}, \vec{x}^d, \vec{\theta})}{\partial\vec{\theta}} \right] \nonumber \\
&\quad= \sum_{d = 1}^{D} \left[ - \frac{\partial E(\vec{y}^d, \vec{x}^d, \vec{\theta})}{\partial\vec{\theta}} + \sum_{\vec{y} \in {\cal Y}} p(\vec{y} | \vec{x}^d, \vec{\theta}) \frac{\partial E(\vec{y}, \vec{x}^d, \vec{\theta})}{\partial\vec{\theta}} \right] \nonumber \\
&\quad= \sum_{d = 1}^{D} \left[ - \frac{\partial E(\vec{y}^d, \vec{x}^d, \vec{\theta})}{\partial\vec{\theta}} + \mathbb{E}_{p(\vec{y} | \vec{x}^d, \vec{\theta})} \frac{\partial E(\vec{y}, \vec{x}^d, \vec{\theta})}{\partial\vec{\theta}} \right] \,.
\end{align}
Direct computation of the gradient~\eqref{eq:likelihood_gradient} is infeasible due to an exponential number of possible variable configurations $\vec y$, which must be considered to compute $\mathbb{E}_{p(\vec{y} | \vec{x}^d, \vec{\theta})} \frac{\partial E(\vec{y}, \vec{x}^d, \vec{\theta})}{\partial\vec{\theta}}$. Inspired by \cite{richard2004multiscale}, in our work we employ sampling based approximation of~\eqref{eq:likelihood_gradient} instead, which we detail in Section~\ref{sec:sampling-based-learning}.

\subsubsection{Stochastic Approximation.} 
The stochastic gradient approximation proposed in~\cite{robbins1951stochastic} is a common way to learn parameters of a CNN nowadays. It allows to perform parameter updates for a single randomly selected input observation, or a small subset of observations, instead of computing the update step for the whole training set at once, as the latter can be very costly. 
Assume that the gradient of some function $f(\theta)$ can be represented as follows:
\begin{align}\label{equ:stohastic-E-gradient}
	\nabla_{\theta}f = \mathbb{E}_{p(y | \theta)} \nabla_{\theta} g(y, \theta) \,.
\end{align}
Then under mild technical conditions the following procedure
\begin{align}\label{equ:stohastic-gradient-alg}
	\theta_{i + 1} = \theta_i - \eta_i \nabla_{\theta}g(y', \theta_i), \text{ where } y' \sim p(y | \theta_i)
\end{align}
and $\eta_i$ is a diminishing sequence of step-sizes, converges to a critical point of the function $f(\theta)$. We refer to \cite{robbins1951stochastic,spall2005introduction} for details, for the cases of both convex and non-convex functions $f(\theta)$. 

\section{Stochastic Optimization Based Learning Framework}\label{sec:sampling-based-learning}

\subsubsection{Stochastic Likelihood Maximization.}
Since the value $\frac{\partial E(\vec{y}^d, \vec{x}^d, \vec{\theta})}{\partial\vec{\theta}}$ does not depend on $y$, we can rewrite the gradient~\eqref{eq:likelihood_gradient} as
\begin{equation}\label{eq:likelihood_gradient_regroup}
\hspace{-1pt}
\frac{\partial \sum_{d = 1}^{D} \log p(\vec{y}^d | \vec{x}^d, \vec{\theta})}{\partial \vec{\theta}}  = \sum_{d = 1}^{D} \mathbb{E}_{p(\vec{y} | \vec{x}^d, \vec{\theta})} \left[ - \frac{\partial E(\vec{y}^d, \vec{x}^d, \vec{\theta})}{\partial\vec{\theta}} + \frac{\partial E(\vec{y}, \vec{x}^d, \vec{\theta})}{\partial\vec{\theta}} \right].
\end{equation}
The summation over samples from the training set can be seen as an expectation over a uniform distribution and therefore 
the index $d$ can be seen as drawn from this uniform distribution. According to this observation we can rewrite~\eqref{eq:likelihood_gradient_regroup} as
\begin{equation}\label{eq:likelihood-gradient-double-mean}
\frac{\partial \sum\limits_{d = 1}^{D} \log p(\vec{y}^d | \vec{x}^d, \vec{\theta})}{\partial \vec{\theta}}  = D\cdot\mathbb{E}_{p(\vec{y},d | \vec{x}^d, \vec{\theta})} \left[ - \frac{\partial E(\vec{y}^d, \vec{x}^d, \vec{\theta})}{\partial\vec{\theta}} + \frac{\partial E(\vec{y}, \vec{x}^d, \vec{\theta})}{\partial\vec{\theta}} \right],
\end{equation}
where $p(\vec{y},d | \vec{x}^d, \vec{\theta})=p(d)p(\vec{y}| \vec{x}^d, \vec{\theta})$ and $p(d)=\frac{1}{D}$.
Assume that we can obtain i.i.d.~samples $\vec{y}'$ from $p(\vec{y} | \vec{x}^d, \vec{\theta})$. Then the following iterative procedure converges to a critical point of the likelihood~\eqref{eq:likelihood} according to~\eqref{equ:stohastic-E-gradient} and~\eqref{equ:stohastic-gradient-alg}
\begin{align}\label{eq:gradient_update}
	\vec{\theta}_{i+1} = \vec{\theta}_{i} - \eta_i \left[ - \frac{\partial E(\vec{y}^d, \vec{x}^d, \vec{\theta_i})}{\partial\vec{\theta}} + \frac{\partial E(\vec{y}', \vec{x}^d, \vec{\theta_i})}{\partial\vec{\theta}} \right] \,,
\end{align}
where $d$ is uniformly sampled from $\{1, \ldots, D\}$ and $\vec{y}' \sim p (\vec{y} | \vec{x}^d, \vec{\theta}_i)$. 

Now we turn to the computation of the stochastic gradient $- \frac{\partial E(\vec{y}^d, \vec{x}^d, \vec{\theta})}{\partial\vec{\theta}} + \frac{\partial E(\vec{y}', \vec{x}^d, \vec{\theta})}{\partial\vec{\theta}}$ itself, provided $\vec{y}^d, \vec{y}', \vec{x}^d$ and $\vec{\theta}$ are given. In the {\em overcomplete representation}~\cite{wainwright2008graphical} the energy~\eqref{eq:crf} reads
\begin{align}
E(\vec{y}, \vec{x}, \theta) = \sum_{f \in {\cal F}} \sum_{\hat{\vec{y}}_f \in {\cal Y}_f} \psi_{f}(\hat{\vec{y}}_f, \vec{x}, \vec{\theta}) \cdot \llbracket \vec{y}_f = \hat{\vec{y}}_f \rrbracket \,,
\end{align}
where expression $\llbracket A \rrbracket$ equals $1$ if $A$ is true and $0$ otherwise. Therefore $\frac{\partial E(\vec{y}, \vec{x}, \vec{\theta})}{\partial \psi_{f}(\hat{\vec{y}}_f, \vec{x}, \vec{\theta})} = \llbracket \vec{y}_f = \hat{\vec{y}}_f \rrbracket$. If
the potential $\psi_{f}(\hat{\vec{y}}_f, \vec{x}, \vec{\theta})$ is an output of a CNN, then the value $- \frac{\partial E(\vec{y}^d, \vec{x}^d, \vec{\theta})}{\partial \psi_{f}(\hat{\vec{y}}_f, \vec{x}^{d}, \vec{\theta})} + \frac{\partial E(\vec{y}', \vec{x}^d, \vec{\theta})}{\partial \psi_{f}(\hat{\vec{y}}_f, \vec{x}^{d}, \vec{\theta})} = - \llbracket \vec{y}_f^d = \hat{\vec{y}}_f \rrbracket + \llbracket \vec{y}_{f}' = \hat{\vec{y}}_f \rrbracket$ is the error to propagate to the CNN. During the back-propagation of this error all parameters $\vec{\theta}$ of the CNN are updated. The overall stochastic maximization procedure for the likelihood~\eqref{eq:likelihood} is summarized in Algorithm~\ref{alg:ll_maximization}. The algorithm is fully defined up to sampling from the distribution $p(\vec{y} | \vec{x}^d, \vec{\theta})$ in Step $5$. We discuss different approaches in the next subsection.

\begin{algorithm}
  \caption{Sampling-based maximization of the likelihood \eqref{eq:likelihood}}\label{alg:ll_maximization}
  \begin{algorithmic}[1]
      \State Initialize parameters $\vec{\theta}_0$ of the CNN-CRF model.
      \For{$i = 1$ \text{to} $M$ (\textit{max. number of iterations})}
      	\State Uniformly sample $d$ from $\{1, \ldots, D\}$
      	\State Perform forward pass of the CNN to get $\psi_{f}(\hat{\vec{y}}, \vec{x}^d, \vec{\theta}_{i-1})$ for each $f \in {\cal F}$ and $\hat{\vec{y}}_f \in {\cal Y}_f$
      	\State Sample $\vec{y}'$ from the distribution $p(\vec{y} | \vec{x}^d, \vec{\theta}_{i-1})$ defined by~\eqref{eq:posterior}
      	\State Compute the error $- \llbracket \vec{y}_f^d = \hat{\vec{y}}_f \rrbracket + \llbracket \vec{y}_{f}' = \hat{\vec{y}}_f \rrbracket$ for each $f \in {\cal F}$ and $\hat{\vec{y}}_f \in {\cal Y}_f$
      	\State Back propagate the error through CNN to obtain a gradient $\nabla_{\vec{\theta}}$
      	\State Update the parameters $\vec{\theta}_i := \vec{\theta}_{i-1} - \eta_i \nabla_{\vec{\theta}}$ 
      \EndFor
      \State \Return{$\vec{\theta}_{M}$}
  \end{algorithmic}
\end{algorithm}

\subsubsection{Sampling.}
Obtaining an exact sample from $p(\vec{y} | \vec{x}, \vec{\theta})$ is a difficult problem for a general CRF due to the exponential size of the set ${\cal Y}\ni\vec{y}$ of all possible configurations. There are, however, ways to mitigate it. The full Markov Chain Monte Carlo (MCMC) sampling method~\cite{geyer1992practical} starts from an arbitrary variable configuration $\vec{y} \in {\cal Y}$ and generates the next one $\vec{y}'$.
In our case this generation can be done with e.g.~Gibbs sampling \cite{Gemans}, as presented in Algorithm~\ref{alg:gibbs}. Algorithm~\ref{alg:gibbs} passes over all variables $y_n$ and updates each of them according to the conditional distribution $p(y_n | \vec{y}_{\setminus n}, \vec{x}, \vec{\theta})$, where ${\setminus n}$ denotes all variable indexes except $n$. 
Let $\mbox{nb}(n)=\{ k\in I | \exists f\in{\cal F}\colon n,k\in f\}$ denote all neighbors of the variable $n$. Note, that due to the Markov property of CRFs~\cite{lauritzen1996}, it holds 
\begin{equation}\label{equ:local-distribution-to-generate}
 p(y_n | \vec{y}_{\setminus n}, \vec{x}, \vec{\theta})=p(y_n | \vec{y}_{\mbox{nb}(n)}, \vec{x}, \vec{\theta})\propto\exp\left(-\sum_{f \in {\cal F}: n \in f} \psi_f(\vec{y}_f, \vec{x}, \vec{\theta})\right)\,.
\end{equation}
Therefore, sampling from this distribution can be done efficiently, since it requires evaluating only those potentials $\psi_f(\vec{y}_f, \vec{x}, \vec{\theta})$ which are dependent on the variable $y_n$, i.e.~for $f\in{\cal F}$ such that $n\in f$. Algorithm~\ref{alg:gibbs} summarizes one iteration of the sampling procedure. Note that it is highly parallelizable~\cite{gonzalez2011parallel} and allows for efficient GPU implementations.
Under mild technical conditions the MCMC sampling process converges to a stationary distribution after a finite number of iterations~\cite{geyer1992practical}. This distribution coincides with  $p(\vec{y} | \vec{x}, \vec{\theta})$. However, such a sampling is time-consuming, because convergence to the stationary distribution may require many iterations and must be performed after each update of the parameters $\vec\theta$.

To overcome this difficulty a contrastive-divergence (CD) method was proposed in~\cite{hinton2002cd} and theoretically justified in~\cite{yuille2004convergence}. For a randomly generated index $d\in\{1,\dots,D\}$ of the training sample one performs a single step of the MCMC procedure starting from a ground-truth variable configuration, which in our case boils down to a single run of Algorithm~\ref{alg:gibbs} for $\vec{y}=\vec{y}^d$. Unfortunately, the sufficient conditions needed to justify this method according to~\cite{yuille2004convergence} do not hold for CRFs in general. Nevertheless, we provide an experimental evaluation of this method in Section~\ref{sec:experiments} along with a different technique described next.

Persistent contrastive divergence (PCD)~\cite{tieleman2008training} is a further development of contrastive divergence, where one step of the MCMC method is performed starting from the sample obtained on a previous learning iteration. It is based on the assumption that the distribution $p(\vec{y}, \vec{x}, \vec{\theta})$ changes slowly from iteration to iteration and a sample from $p(\vec{y} | \vec{x}^d, \vec{\theta}_{i-1})$ is close enough to a sample from $p(\vec{y} | \vec{x}^d, \vec{\theta}_{i})$.
Moreover, when getting closer to a critical point, the gradient becomes smaller and therefore $p(\vec{y}, \vec{x}, \vec{\theta}_{i})$ deviates less from $p(\vec{y}, \vec{x}, \vec{\theta}_{i-1})$. Therefore, close to a critical point the generated samples can be seen as samples from the stationary distribution of the full MCMC method, which coincides with the desired one $p(\vec{y} | \vec{x}, \vec{\theta})$.

With the above description of the possible sampling procedures the whole joint CNN-CRF learning Algorithm~\ref{alg:ll_maximization} is well-defined.

\begin{algorithm}
  \caption{Gibbs sampling}\label{alg:gibbs}
  \begin{algorithmic}[1]
    \Require{A variable configuration $\vec{y}\in{\cal Y}$}  
      	\For{n = 1, \ldots, N}
      		\State $y_n'$ is sampled from $p(y_n | \vec{y}_{\setminus n}, \vec{x}, \vec{\theta})$
      		\State $y_n \gets y_n'$
      	\EndFor
      \State \Return{$\vec{y}$}
  \end{algorithmic}
\end{algorithm}

\section{Comparison to Alternative Approaches}~\label{sec:competing-approaches} 

\subsubsection{Unrolled Inference.} In contrast to the learning method with the unrolled inference proposed in \cite{ZhengJRVSDHT15} and \cite{schwing/arxiv2015}, our approach is not limited to Gaussian pairwise potentials. In our training procedure the potentials $\phi_f(\vec{y}_f, \vec{x}, \vec{\theta})$ can have arbitrary form.

\subsubsection{The piece-wise training} method~\cite{LinSRH15} is able to handle arbitrary potentials in CRFs. However, maximization of the likelihood~\eqref{eq:likelihood} in that work is substituted with  
\begin{align}
	(\arg)\max_{\vec{\theta}} \sum_{d = 1}^{D} \sum_{f \in {\cal F}} \Bigl[-\psi_f(\vec{y}_f^d, \vec{x}^d, \vec{\theta})-
	\log\sum_{\vec{y}_f\in{\cal Y}_f}\exp (-\psi_f(\vec{y}_f, \vec{x}^d, \vec{\theta}))\Bigr]\,,
\end{align}
which lacks a sound theoretical justification.

\subsubsection{LP-relaxation and fractional entropy based approximation} is employed in~\cite{ChenSYU15}. 
As mentioned above, there is no clear evidence that the fractional entropy always leads to tight likelihood approximations. Additionally, the method requires a lot of memory: to avoid the time consuming, full message passing based inference, the gradient steps w.r.t.~the CNN parameters $\vec\theta$ and minimization over the dual variables of the LP-relaxation of the CRF are interleaved with each other. This requires to store current values of the dual variables for {\em each} training sample. The number of dual variables is proportional to the number of labels used in the CRF as well as to the number of its factors. So, for example in our experiments we use a dataset containing 2000 images of the approximate size $320 \times 120$. The corresponding CRF has $20$ labels and around $10^6$ pairwise factors (see Section~\ref{sec:experiments} for details). The dual variables stored by the method~\cite{ChenSYU15} would require around $200$MB per image and $0.4$TB for the whole dataset.
Note that our approach requires to store only the current variable configuration $\vec{y}$ for each of the $D$ training samples, when used with the PCD sampling. 
Therefore, it requires only $78$MB of working storage for the whole dataset. The difference between our method and the method proposed in~\cite{ChenSYU15} gets even more pronounced for larger problems and datasets, such as the augmented Pascal VOC dataset~\cite{pascal-voc-2012,BharathICCV2011} containing $10000$ images with $500 \times 300$ pixels each.


\section{Experiments}
\label{sec:experiments}

In the experimental evaluation we consider the problem of semantic body-parts segmentation from a single depth image \cite{Denil2013a}. We specify a CRF, which has unary potentials dependent on a CNN. We test different sampling options in Algorithm~\ref{alg:ll_maximization} and compare our approach with another CRF-CNN learning framework proposed in~\cite{ZhengJRVSDHT15}. Additionally, we analyze the trained model, in order to understand whether it can capture an object shape and contextual information.

\subsubsection{Dataset and evaluation.} We apply our approach to the challenging task of predicting human body parts from a depth image. To the best of our knowledge, there is no publicly available dataset for this task that contains real depth images. For this reason, in \cite{Denil2013a}, a set of synthetically rendered depth images along with the corresponding ground truth labelings were introduced (see examples in Fig.~\ref{fig:kinect} (left column)). In total there are $19$ different body part labels, and one additional label for the background. The dataset is split into 2000 images for training and 500 images for testing. As a quality measure, the authors use the averaged per-pixel accuracy for body parts labeling, excluding the background. This makes sense since the background can be easily identified from the depth map.

\subsubsection{Our model.} 
\begin{figure*}
    		\begin{subfigure}[b]{0.32\textwidth}
        		\includegraphics[width=0.9\linewidth]{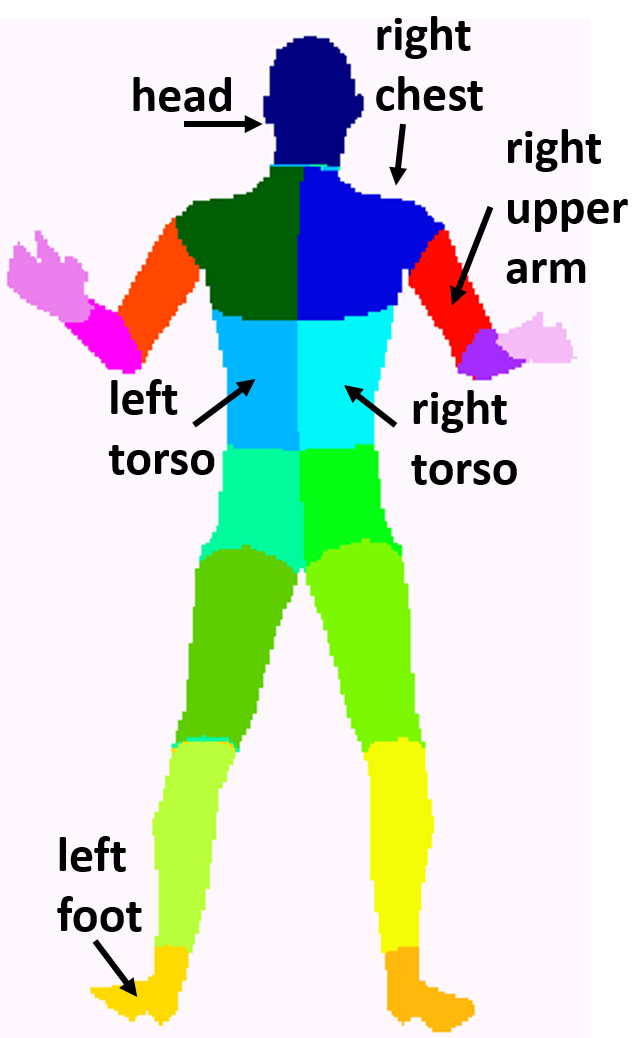}
        		\subcaption{}
        		\label{fig:pw}
        	\end{subfigure}
        	\begin{subfigure}[b]{0.30\textwidth}
        			\begin{subfigure}[b]{1.0\textwidth}
		        		\includegraphics[width=0.7\linewidth]{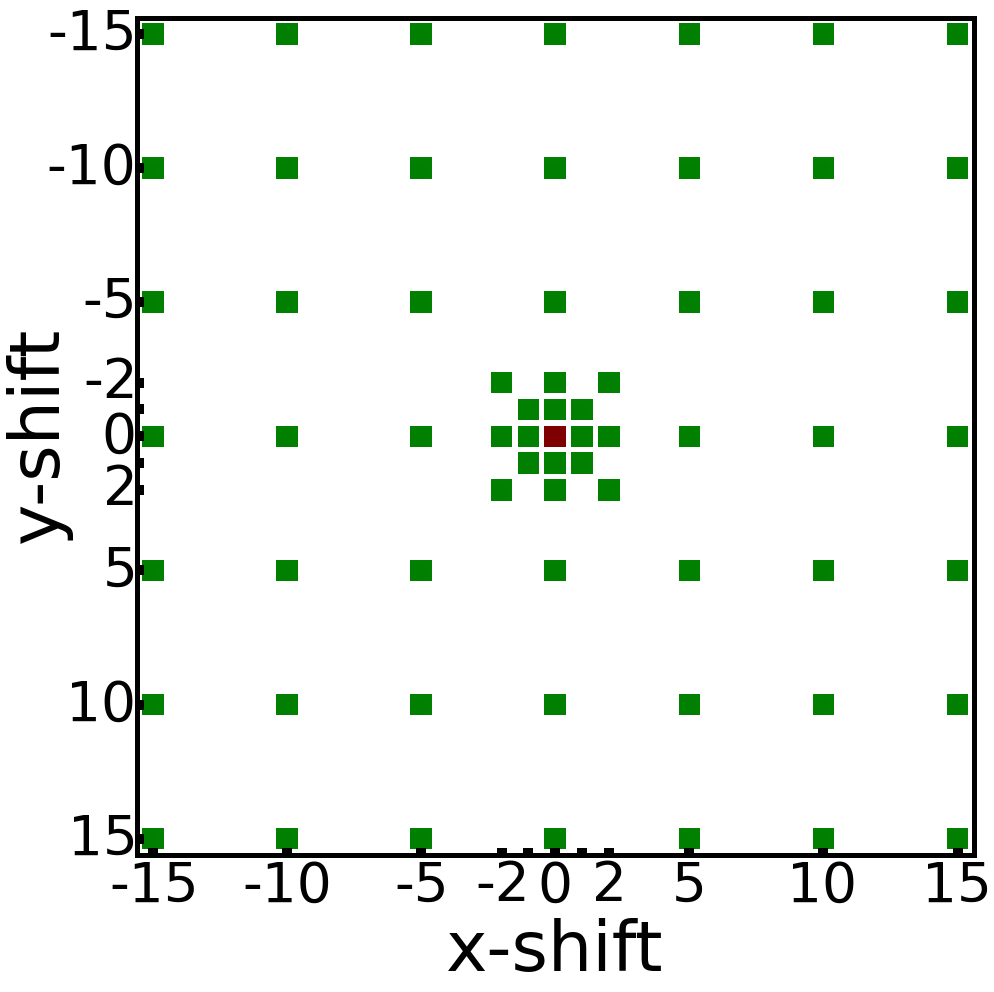}
		        		\subcaption{}
		        		\label{fig:pw0}
		        	\end{subfigure}
		        	\begin{subfigure}[b]{1.0\textwidth}
		        		\includegraphics[width=0.9\linewidth]{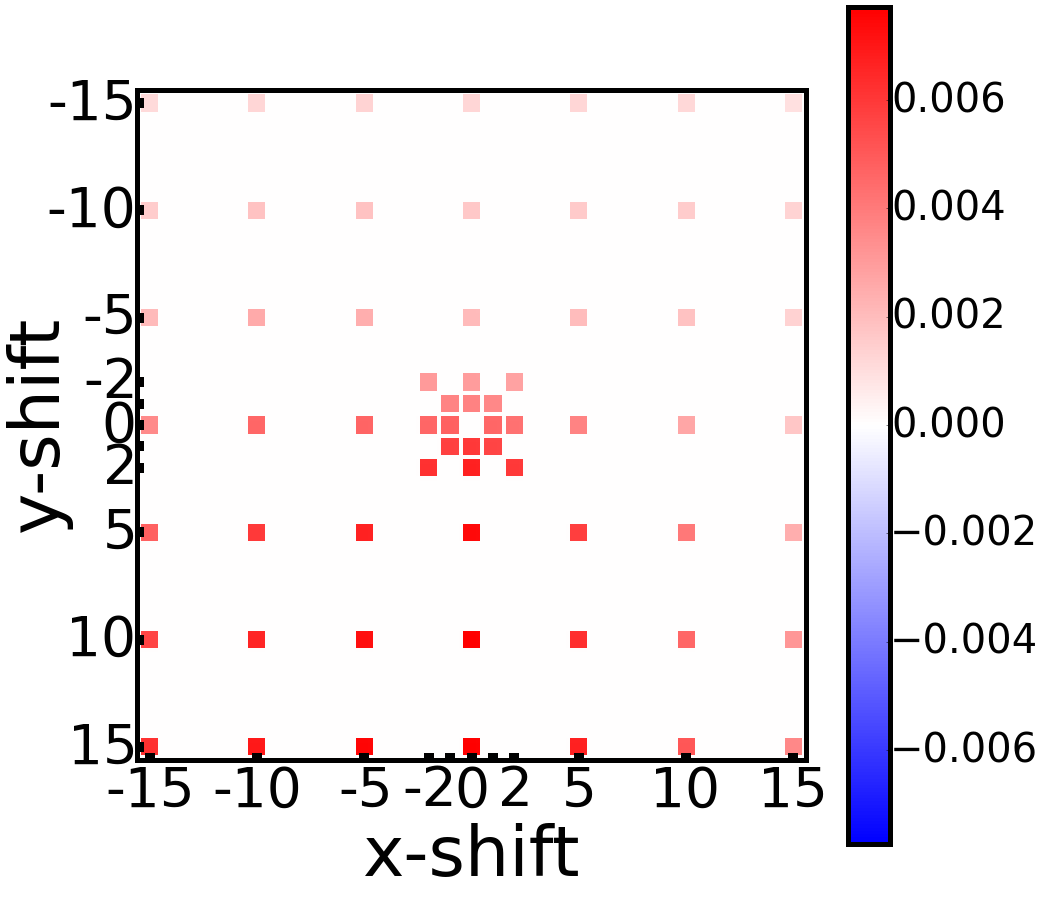}
		        		\subcaption{}
		        		\label{fig:pw_1}
		        	\end{subfigure}
        	\end{subfigure}
        	\begin{subfigure}[b]{0.30\textwidth}
		        	\begin{subfigure}[b]{1.0\textwidth}
		        		\includegraphics[width=0.9\linewidth]{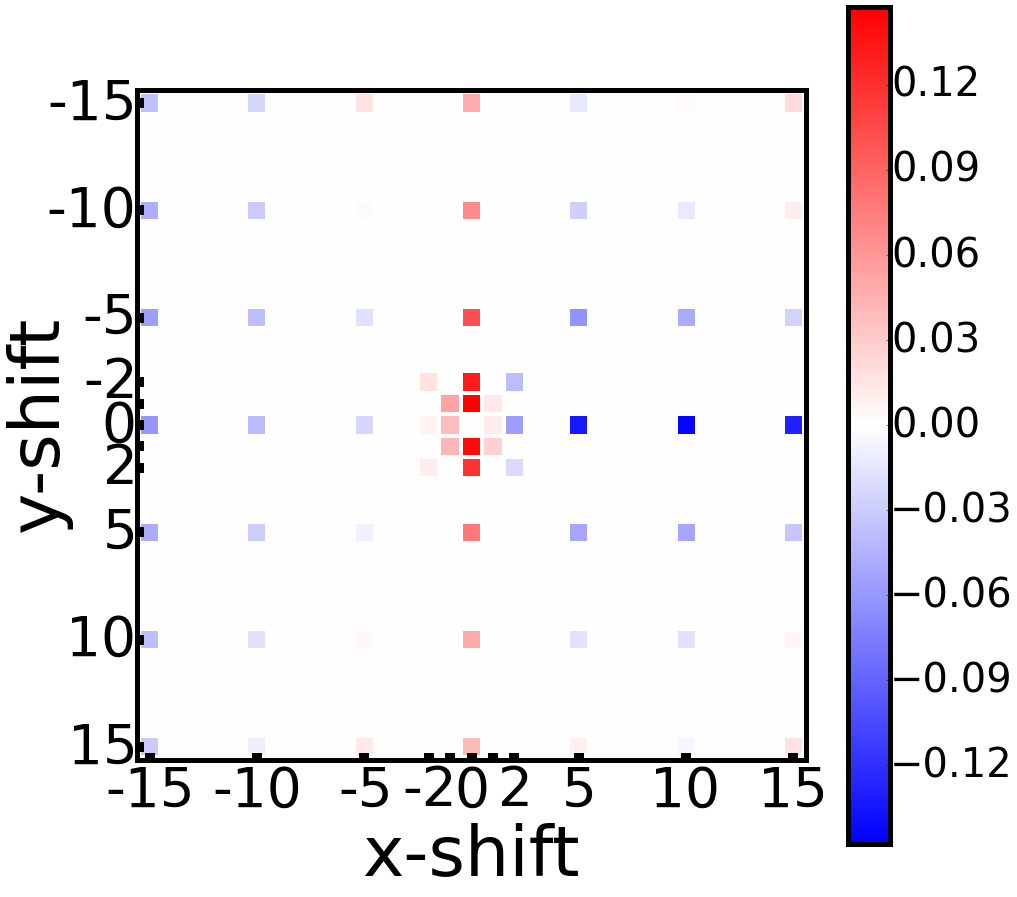}
		        		\subcaption{}
		        		\label{fig:pw_2}
		        	\end{subfigure}
		        	\begin{subfigure}[b]{1.0\textwidth}
		        		\includegraphics[width=0.9\linewidth]{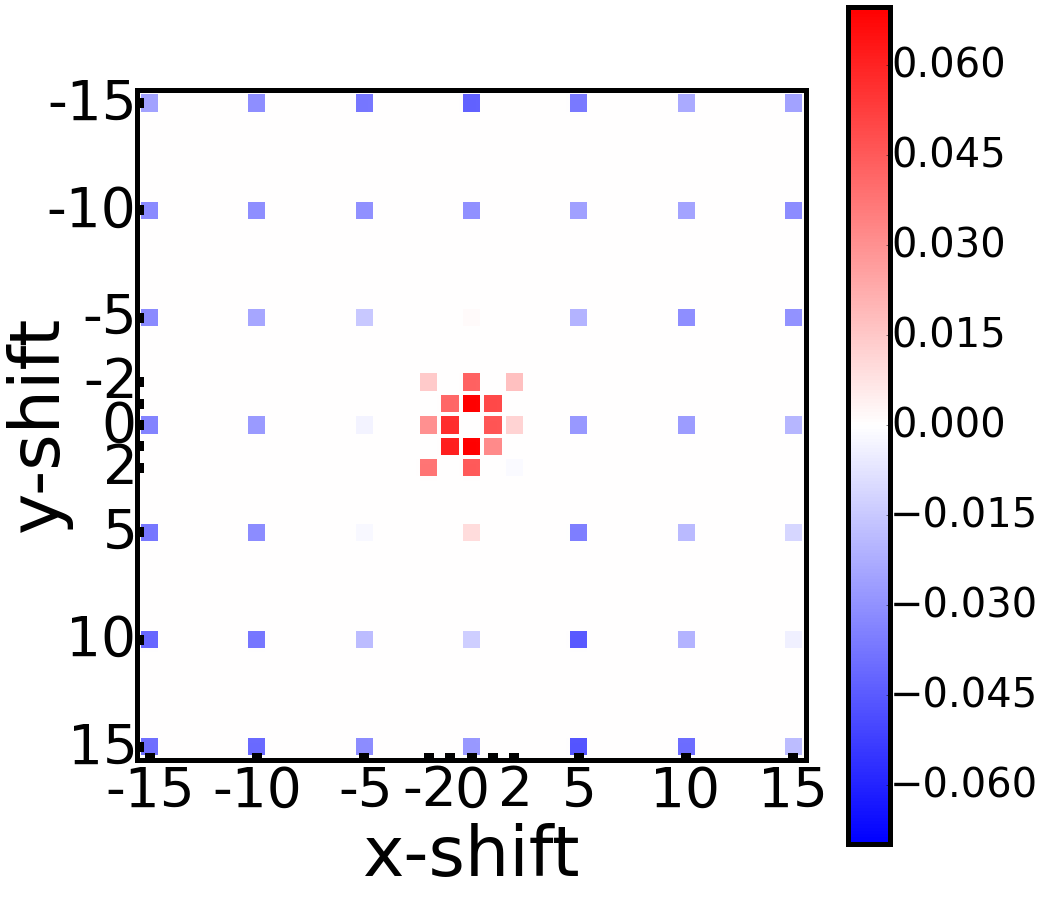}
		        		\subcaption{}
		        		\label{fig:pw_3}
		        	\end{subfigure}
		\end{subfigure}
        \caption{{\bf Model Insights.} (a) Illustrating the $19$ body parts of a human. (c-e)~Weights of pairwise factors for different pairs of labels, see details below. (b)~Neighborhood structure for pairwise factors. The center pixel (red) is connected via pairwise factors to all green pixels. Note that ``opposite'' edges share same weights, e.g.~the edge with $x,y$-shift ($5, 10$) has the same weights as the edge with $x,y$-shift ($-5,-10$). (c) Weights for pairwise potentials that connect the label ``head'' with the label ``foot''. Red means a high energy value, i.e.~a discouraged configuration, while blue means the opposite. Since there is no sample in the training dataset where a foot is close to a head, all edges are positive or close to 0. Note that the zero weights can occur even for very unlikely configurations. The reason is that during training these unlikely configurations did not occur. (d) Weights for pairwise potentials that connect the label ``left torso'' with the label ``right torso''. The potentials enforce a straight, vertical border between the two labels, i.e.~there is a large penalty for ``left torso'' on top (or below) of ``right torso'' (x-shift 0, y-shift arbitrary). Also, it is encouraged that ``right torso'' is to the right of the ``left torso''  (Positive x-shift and y-shift 0). (e)~Weights for pairwise potentials that connect the label ``right chest'' with the label ``right upper arm''. It is discouraged that the ``right upper arm'' appears close to ``right chest'', but this configuration can occur at a certain distance. Since the training images have no preferred  arm-chest configurations, all directions have similar weights.}
        \label{fig:insight1}
    \end{figure*}

Following \cite{nowozin2011decision}, in our experiments, we use a pixel-level CRF that is able to capture geometrical layout and context. The state vector $\vec{y}$ defines a per pixel labeling. Therefore the number $N$ of coordinates in $\vec y$ is equal to the number of pixels in a depth image, which has dimensions varying around $130\times 330$. For all $n \in \{1, \ldots, N\}$ the label space is ${\cal Y}_n = \{1, \ldots, 20\}$. The observation $\vec{x}$ represents a depth image. Our CRF has the following energy function $E(\vec{y}, \vec{x}, \vec{\theta})$:
\begin{align}\label{eq:our_crf}
	E(\vec{y}, \vec{x}, \vec{\theta}) = \sum_{n = 1}^{N} \psi_n(y_n, \vec{x}, \vec{\theta}) + \sum_{c \in {\cal C}} \sum_{(i,j) \in E_c} \psi_{c}(y_i, y_j, \vec{\theta}) \,,
\end{align}
where $\psi_n(y_n, \vec{x}, \vec{\theta})$ are unary potentials that depend on a CNN. Our CRF has $|{\cal C}|$ classes of pairwise potentials. All potentials of one class are represented by a learned value table, which they share. The neighborhood structure of the CRF is visualized in Fig. \ref{fig:pw0}. All pixels are connected to $64$ neighbors, apart from those close to the image border. 

The local distribution~\eqref{equ:local-distribution-to-generate} used by the sampling Algorithm~\ref{alg:gibbs} takes the form:
\begin{align}\label{eq:gs}
p_i(y_i{=}l|x,y_{R\setminus i};\theta)\propto\exp\Bigl[-\psi_i(l)-\sum_c\bigl(\psi_c(l,y_{j'})+\psi_c(y_{j''},l)\bigr)\Bigr] .
\end{align}
Note that according to our CRF architecture there are exactly two edges (apart from the nodes close to the image border) in each edge class $c$ that are incident to a given node $i$. The corresponding neighboring nodes are denoted by $j'$ and $j''$ in~\eqref{eq:gs}. 

As mentioned above, the unary terms of our CRF model depend on the image via a CNN. Since most existing pre-trained CNNs~\cite{long2014fully,NIPS2012_4824,Simonyan14c} use RGB images as input, for the depth input we use our own fully convolutional architecture and train it from scratch. Moreover, since some body parts, such as hands, are relatively small, we use the architecture that does not reduce the resolution in intermediate layers.  This allows us to capture fine details. All intermediate layers have $50$ output channels and a stride of one. The final layer has $20$ output channels that correspond to the output labels. The architecture of our CNN is summarized in Table~\ref{tab:CNN}. During training, we optimize the cross-entropy loss. The CNN is trained using stochastic gradient descent with the momentum $0.99$ and with the batch size $1$.\footnote{We use the commonly adopted terminology from the CNN literature for technical details, to allow reproducibility of our results.}

\begin{table*}[t]
\tabcolsep=0.05cm
\center
\caption{CNN archutecture for body parts segmentation.}\label{tab:CNN}
\begin{tabular}{c|c|c|c|c|c|c|c|c|c}
\toprule
Layer  & conv1          & relu1 & conv2          & relu2 & conv3          & maxpool1     & relu3 & conv4        & Softmax \\

Kernel size & $41 \times 41$ &   -    & $17 \times 17$ &   -    & $11 \times 11$ & $3 \times 3$ &   -    & $5 \times 5$ &   -    \\

Output channels & $50$ &   $50$  & $50$ &   $50$    & $50$ & $50$ &   $50$  & $20$ &   $20$    \\
\bottomrule
\end{tabular}
\end{table*}

In our experiments, we consider two learning scenarios: {\em separate} learning and {\em joint} (end-to-end) learning. In both cases we start the learning procedure from the same pre-trained CNN. For separate learning only the CRF parameters (pairwise potentials) are updated, whereas the CNN weights (unary potentials of the CRF) are kept fixed. In contrast, for joint (end-to-end) learning all parameters are updated. During the test-time inference we empirically observed that starting Gibbs sampling (Algorithm~\ref{alg:gibbs}) from a random labeling can lead to extremely long runtimes. To speed-up the burn-in-phase, we use the marginal distribution of the CNN without CRF. This means that the first sample is drawn from the marginal distribution of the pre-trained CNN.

We also experiment with different sampling strategies during the training phase: we considered (i) the contrastive-divergence with $K$ sampling iterations, denoted as {\tt CD-K} for $K$ equal to $1$, $2$, $5$ and (ii) the persistent contrastive-divergence  {\tt PCD}.  

\subsubsection{Baselines.} We compare our approach to the method of~\cite{Denil2013a}, which introduced this dataset. Their approach is based on a random forest model. Unfortunately, we were not able to compare to the recent work \cite{shaoqing15grrf}, which extends \cite{Denil2013a}, and is also based on random forests. The reason is that in the work~\cite{shaoqing15grrf} its own evaluation measure is used, meaning that the accuracy of only a small subset of pixels is evaluated. This subset is chosen in such a way that each of the $20$ classes is represented by the same number of pixels. We are concerned, however, that such small pixel subsets may introduce a bias. Furthermore, we did not have this subset at our disposal. Since our main aim is to evaluate CNN-based CRF models, we compare to the approach \cite{ZhengJRVSDHT15}. As described above, they incorporate a densely connected Gaussian CRF model into the CNN as a Recurrent Neuronal Network of the corresponding Mean Field inference steps. This approach has recently been the state-of-the-art in the VOC2012 object segmentation challenge.

\begin{table*}[t]
\tabcolsep=0.8cm
\center
\caption{Average per-pixel accuracy for all foreground parts. {\em Separate} learning means that weights of the respective CNN were trained prior to CRF parameters. In contrast, {\em joint} training means that all weights were learned jointly, starting with a pre-trained CNN. We obverse that joint training is superior to separate training, and furthermore that the model of~\cite{ZhengJRVSDHT15}, which is based on a dense Gaussian CRF, is inferior to our generic CRF model.}\label{tabel:results}
\begin{tabular}{c|c|c}
Method & Learning &  Accuracy \\
\midrule
Online Random Forest \cite{Denil2013a} & - &$\approx 79.0\%$ \\
CNN  & - & $84.47$ \\
CNN + CRF \cite{ZhengJRVSDHT15} & separate  & $86.55\%$ \\
CNN + CRF \cite{ZhengJRVSDHT15} & joint & $88.17\%$ \\
CNN + CRF (ours) {\tt PCD}  & separate & $87.62\%$ \\
CNN + CRF (ours) {\tt CD-1} & joint & 88.17\% \\
CNN + CRF (ours) {\tt CD-2} & joint & 88.15\% \\
CNN + CRF (ours) {\tt CD-5} & joint & 88.23\% \\
CNN + CRF (ours) {\tt PCD} & joint &  \bf{89.01\%} \\
\bottomrule
\end{tabular}
\end{table*}

 		\begin{figure*}
    		\centering
        \begin{subfigure}[b]{0.9\textwidth}
	        	\begin{subfigure}[b]{0.16\textwidth}
	        \center
	        \caption*{\center Input Depth}
	            \includegraphics[width=1.0\linewidth]{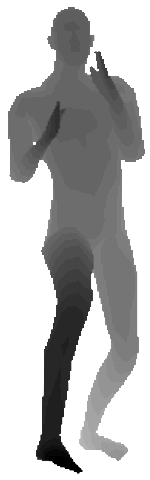}
	        \caption*{ }
	        \end{subfigure}
	        \begin{subfigure}[b]{0.16\textwidth}
	        \center
	        \caption*{\center Ground Truth}
	            \includegraphics[width=1.0\linewidth]{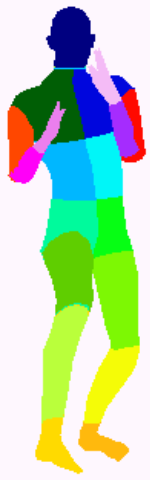}
	        \caption*{ }
	        \end{subfigure}
	        \begin{subfigure}[b]{0.16\textwidth}
	        \center
	        \caption*{\center CNN}
	            \includegraphics[width=1.0\linewidth]{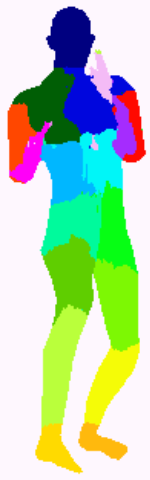}
	        \caption*{86.53\%}
	        \end{subfigure}
	        \begin{subfigure}[b]{0.16\textwidth}
	        \center
	        \caption*{\center \cite{ZhengJRVSDHT15} joint learning}
	            \includegraphics[width=1.0\linewidth]{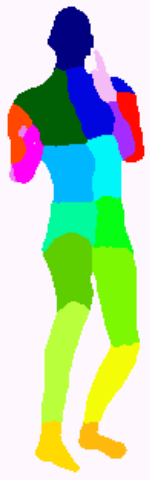}
	        \caption*{90.88\%}
	        \end{subfigure}
	        	\begin{subfigure}[b]{0.16\textwidth}
	        \center
	        \caption*{\center Ours separate learning}
	            \includegraphics[width=1.0\linewidth]{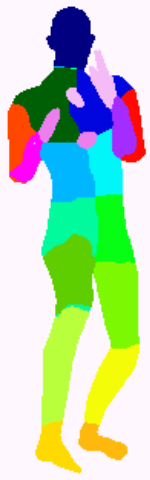}
	        \caption*{90.38\%}
	        \end{subfigure}
	        	\begin{subfigure}[b]{0.16\textwidth}
	        \center
	        \caption*{\center Ours joint learning}
	            \includegraphics[width=1.0\linewidth]{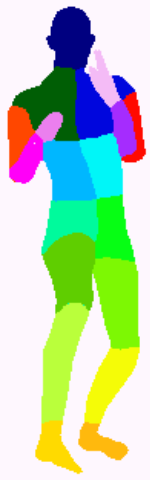}
	        \caption*{93.14\%}
	        \end{subfigure}
        \end{subfigure}
        \begin{subfigure}[b]{0.9\textwidth}
	        	\begin{subfigure}[b]{0.16\textwidth}
	        \center
	            \includegraphics[width=1.0\linewidth]{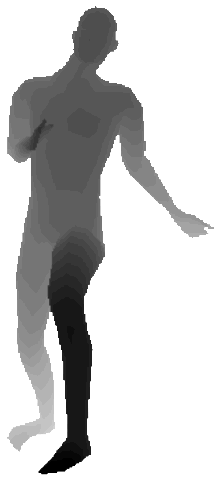}
	        \caption*{ }
	        \end{subfigure}
	        \begin{subfigure}[b]{0.16\textwidth}
	        \center
	            \includegraphics[width=1.0\linewidth]{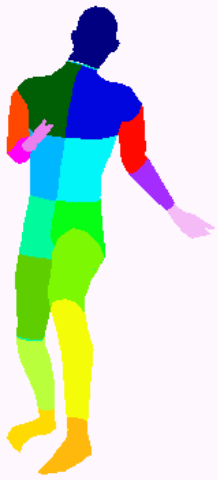}
	        \caption*{ }
	        \end{subfigure}
	        \begin{subfigure}[b]{0.16\textwidth}
	        \center
	            \includegraphics[width=1.0\linewidth]{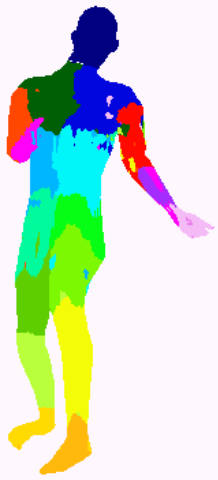}
	        \caption*{83.93\%}
	        \end{subfigure}
	        \begin{subfigure}[b]{0.16\textwidth}
	        \center
	            \includegraphics[width=1.0\linewidth]{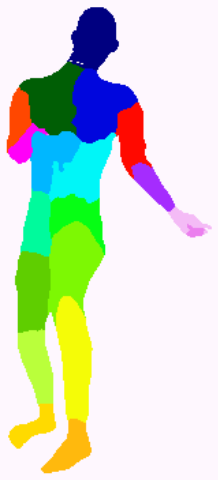}
	        \caption*{88.20\%}
	        \end{subfigure}
	        	\begin{subfigure}[b]{0.16\textwidth}
	        \center
	            \includegraphics[width=1.0\linewidth]{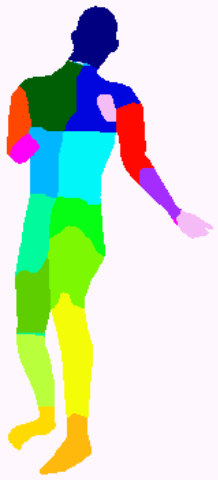}
	        \caption*{87.74\%}
	        \end{subfigure}
	        	\begin{subfigure}[b]{0.16\textwidth}
	        \center
	            \includegraphics[width=1.0\linewidth]{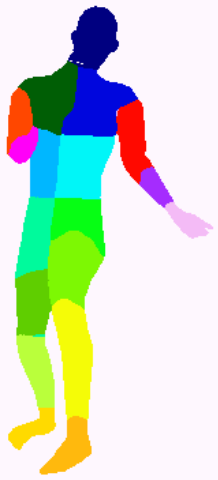}
	        \caption*{90.27\%}
	        \end{subfigure}
        \end{subfigure}
        \begin{subfigure}[b]{0.9\textwidth}
	        	\begin{subfigure}[b]{0.16\textwidth}
	        \center
	            \includegraphics[width=1.0\linewidth]{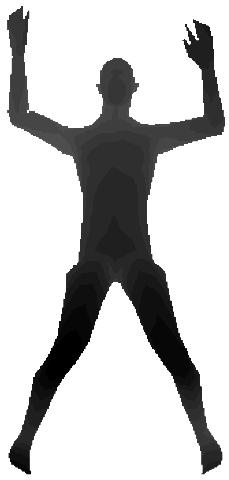}
	        \caption*{ }
	        \end{subfigure}
	        \begin{subfigure}[b]{0.16\textwidth}
	        \center
	            \includegraphics[width=1.0\linewidth]{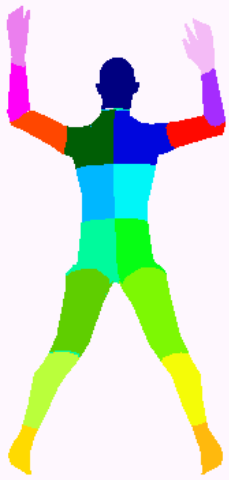}
	        \caption*{ }
	        \end{subfigure}
	        \begin{subfigure}[b]{0.16\textwidth}
	        \center
	            \includegraphics[width=1.0\linewidth]{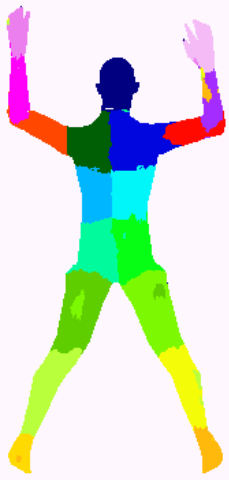}
	        \caption*{90.15\%}
	        \end{subfigure}
	        \begin{subfigure}[b]{0.16\textwidth}
	        \center
	            \includegraphics[width=1.0\linewidth]{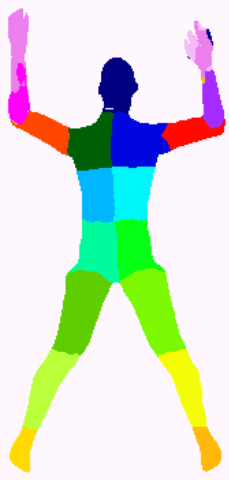}
	        \caption*{92.62\%}
	        \end{subfigure}
	        	\begin{subfigure}[b]{0.16\textwidth}
	        \center
	            \includegraphics[width=1.0\linewidth]{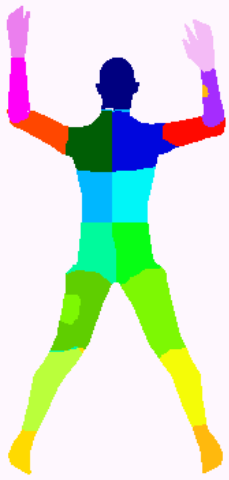}
	        \caption*{91.98\%}
	        \end{subfigure}
	        	\begin{subfigure}[b]{0.16\textwidth}
	        \center
	            \includegraphics[width=1.0\linewidth]{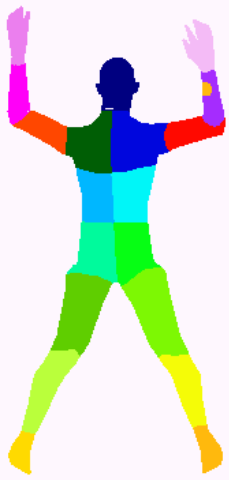}
	        \caption*{93.58\%}
	        \end{subfigure}
        \end{subfigure}
        \caption{{\bf Results.} (From left to right). The input depth image. The corresponding ground truth labeling for all body parts. The result of a trained CNN model. The result of \cite{ZhengJRVSDHT15} using an end-to-end training procedure. Our results with separate learning and joint learning, respectively. Below each result we give the averaged pixel-wise accuracy for all body parts.}
        \label{fig:kinect}
    \end{figure*}

    	\begin{figure*}
        	\begin{subfigure}[b]{0.245\textwidth}
	        	\includegraphics[width=0.9\linewidth]{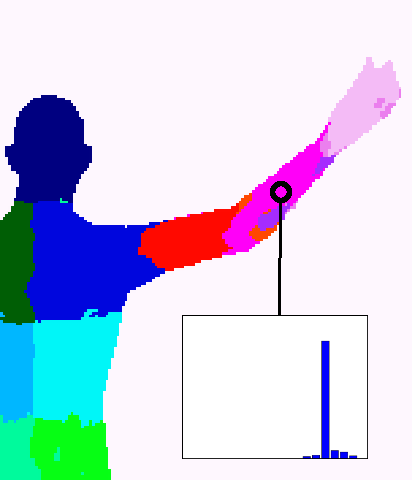}
	        	\subcaption{}
	        	\label{fig:unary_example_unary_pretrain}
        	\end{subfigure}
        	\begin{subfigure}[b]{0.245\textwidth}
	        	\includegraphics[width=0.9\linewidth]{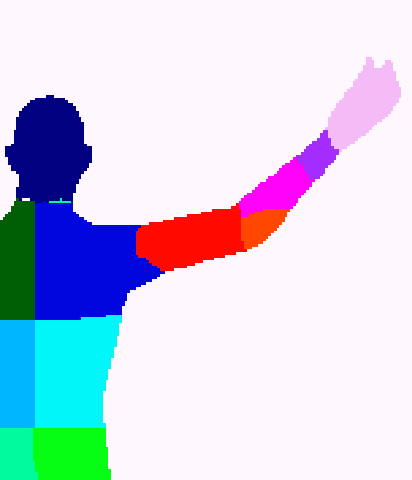}
	        	\subcaption{}
	        	\label{fig:unary_example_unary_joint}
		\end{subfigure}
		\begin{subfigure}[b]{0.245\textwidth}
	        	\includegraphics[width=0.9\linewidth]{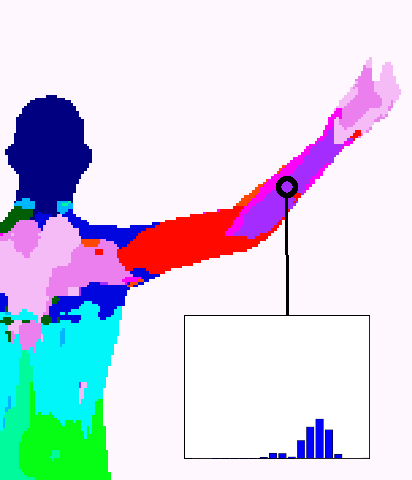}
	        	\subcaption{}
	        \label{fig:unary_example_crf_separate}
		\end{subfigure}
		\begin{subfigure}[b]{0.245\textwidth}
	        	\includegraphics[width=0.9\linewidth]{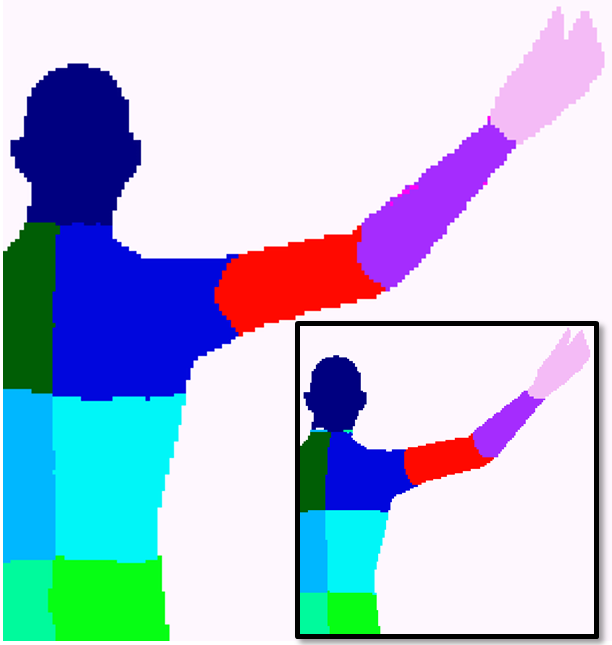}
	        	\subcaption{}
	        	\label{fig:unary_example_crf_joint}
		\end{subfigure}
        \caption{{\bf Model Insights.} (a) The most likely labeling for a separately trained CNN. For the circled pixel, the local marginal distribution is shown. (b) Max. marginal labeling of a separately trained CRF, which uses the CNN unaries from (a), i.e.~our approach with separate learning. We observe that unaries are spatially smoothed-out. (c) Most likely labeling of a CNN that was jointly trained with the CRF. The labeling looks worse than (a). However, the main observation is that the pixel-wise marginal distributions are more ambiguous than in (a), see the circled pixel. (d) The final, max-marginal labeling of the jointly trained CRF model, which is considerably better than the result in (b). The reason is that due to the ambiguity in the local unary marginals, the CRF has more power to find the correct body part configuration. The inlet shows the ground truth labeling.}
        \label{fig:insight2}
    \end{figure*}

\subsubsection{Results.} 
Qualitative and quantitative results are shown in Fig.~\ref{fig:kinect} and Table~\ref{tabel:results} respectively. Our method with joint learning is performing best. In particular, the persistent contrastive-divergence version shows the best results, which conforms to the observations made in other works~\cite{tieleman2008training}. The CNN-CRF approach of \cite{ZhengJRVSDHT15} is inferior to ours. Note that the accuracy difference of $1\%$ can mean that e.g.~a complete hand is incorrectly labeled. We attribute this to the fact that for this task the spatial layout of body parts is of particular importance. The underlying dense Gaussian CRF model of \cite{ZhengJRVSDHT15} is rotational invariant and cannot capture contextual information such as ``the head has to be above the torso''. Our approach is able to capture this, which we explain in detail in Fig.~\ref{fig:insight1} and~\ref{fig:insight2}. We expect that even higher levels of accuracy can be achieved by exploring different network designs and learning strategies, which we leave for future work.

\section{Discussion and Future Work}
We have presented a generic CRF model where a CNN models unary factors. We have introduced an efficient and scalable maximum likelihood learning procedure to train all model parameters jointly. By doing so, we were able to train and test on large-size factor graphs. We have demonstrated a performance gain over competing techniques for semantic labeling of body parts. We have observed that our generic CRF model can capture the shape and context information of relating body parts.

There are many exciting avenues for future research. We plan to apply our method to other application scenarios, such as semantic segmentation of RGB images. In this context, it would be interesting to combine the dense CRF model of \cite{ZhengJRVSDHT15} with our generic CRF model. Note that a dense CRF is implicitly modeling the property that objects have a compact color distribution, see \cite{cgf/Cheng15}, which is a complementary modeling power to our generic CRF model.

\subsubsection{Acknowledgements.} This work was supported by: European Research Council (ERC) under the European Union's Horizon 2020 research and innovation programme (grant agreement No 647769); German Federal Ministry of Education and Research (BMBF, 01IS14014A-D); EPSRC EP/I001107/2; ERC grant ERC- 2012-AdG 321162-HELIOS. The computations were performed on an HPC Cluster at the Center for Information Services and High Performance Computing (ZIH) at TU Dresden.

The final publication is available at {\tt link.springer.com}.

\bibliographystyle{splncs}
\bibliography{0142}



\end{document}